%% file: svrhm_2022.tex
\documentclass{article}

\usepackage[preprint]{svrhm_2022}

\usepackage[utf8]{inputenc} 
\usepackage[T1]{fontenc}    
\usepackage{hyperref}       
\usepackage{url}            
\usepackage{booktabs}       
\usepackage{amsfonts}       
\usepackage{nicefrac}       
\usepackage{microtype}      
\usepackage{xcolor}         
\input{math_commands.tex}   
\usepackage{graphicx}       
\usepackage{dirtytalk}

\title{Stochastic Gradient Descent Captures \\ How Children Learn About Physics}

\author{
Luca M. Schulze Buschoff \thanks{Corresponding author: luca.schulze-buschoff@tuebingen.mpg.de}
\qquad \qquad
Eric Schulz
\qquad \qquad
Marcel Binz
\\ \\
MPRG Computational Principles of Intelligence \\
Max Planck Institute for Biological Cybernetics \\
Tübingen, Germany}

\begin{document}

\maketitle

\begin{abstract}
As children grow older, they develop an intuitive understanding of the physical processes around them. They move along developmental trajectories, which have been mapped out extensively in previous empirical research. We investigate how children's developmental trajectories compare to the learning trajectories of artificial systems. Specifically, we examine the idea that cognitive development results from some form of stochastic optimization procedure. For this purpose, we train a modern generative neural network model using stochastic gradient descent. We then use methods from the developmental psychology literature to probe the physical understanding of this model at different degrees of optimization. We find that the model's learning trajectory captures the developmental trajectories of children, thereby providing support to the idea of development as stochastic optimization.
\end{abstract}

\section{Introduction}

    More than 70 years ago, \citet{turing1950computing} famously suggested that \say{instead of trying to produce a programme to simulate the adult mind, why not rather try to produce one which simulates the child's? If this were then subjected to an appropriate course of education one would obtain the adult brain.} If we want to take Turing’s proposal seriously, we have to ask ourselves: how do children learn?
    
    The physical laws of nature are one of the earliest things that children learn \citep{spelke2007core, lake2017building}. Therefore, they can serve as an ideal testbed for investigating this question. There has already been a substantial amount of work across different research areas to understand and reproduce the human ability for physical reasoning. On the one hand, empirical work in developmental psychology has provided us with a precise understanding of the different developmental stages that children undergo during their cognitive development \citep{baillargeon1996infants, baillargeon2004infants}. On the other hand, machine learning researchers have started to successfully apply tools from deep learning to build models that mimic the intuitive physical understanding of people \citep{battaglia2013simulation, lerer2016learning, zhang2016comparative, piloto2022learn, smith2019modeling, smith2020fine}.
    
    Even though there are a number of models for physical reasoning, they typically focus on reproducing adult-level performance. The goal of the present paper is to instead -- in the spirit of Turing -- compare the learning trajectories of artificial systems to the developmental trajectories of children. We are particularly interested in examining the idea of \emph{development as stochastic optimization}, which states that cognitive development results from some form of stochastic optimization procedure \citep{gopnik2017changes, ullman2020bayesian, giron2022developmental}.
    
    \clearpage
    
    \begin{figure}[!h]
        \centering
        \includegraphics[width=\textwidth]{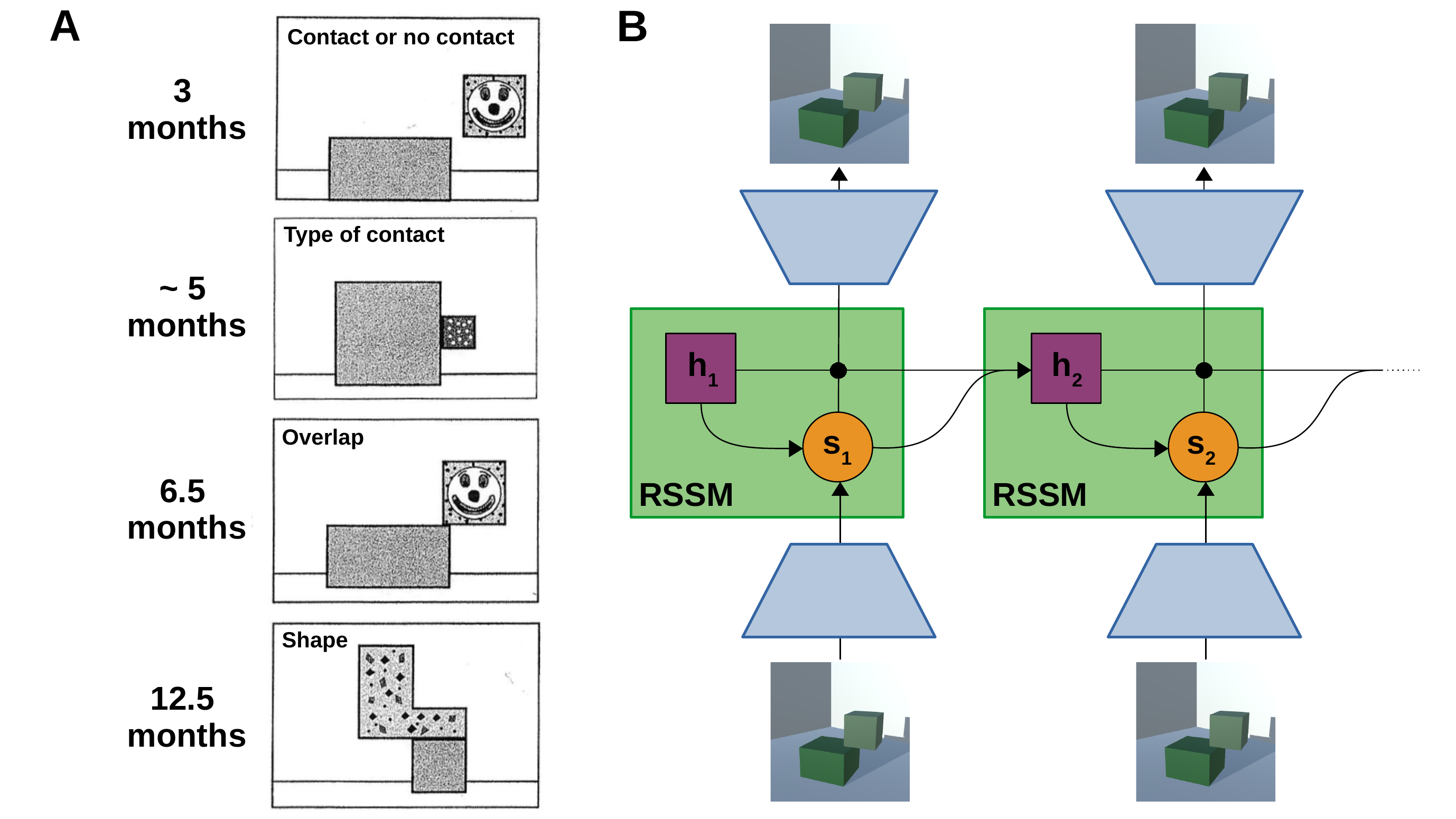}
        \caption{A: Developmental trajectory for support events outlined by \citep{baillargeon1996infants}. The illustrations are taken from \citet{baillargeon1996infants} and they show the physical rules acquired at the respective ages. B: Illustration of our generative video prediction model.}
        \label{fig:model_graph}
    \end{figure}
    
    To test this hypothesis, we train a deep generative model on video sequences on a physical reasoning task. We then probe the knowledge of this model at different training epochs using violation-of-expectation methods \citep{baillargeon2004infants, piloto2018probing, smith2019modeling} and compare it to the knowledge of children at different ages. We find that the acquisition order of concepts in this model aligns with that of children, thereby providing support to the idea of development as stochastic optimization.

\section{Methods}
        
    \subsection{Support events}
    \label{sec:stimuli}
    
        Infants' physical reasoning abilities have been investigated in many different domains. Here, we use support events (such as the configurations of block stacks shown in Figure \ref{fig:model_graph}A) as an exemplary physical reasoning task for comparing the learning trajectories of artificial systems to the developmental trajectories of children. 
        
        \citet{baillargeon1996infants} has shown that, as infants grow older, they make use of increasingly complex rules to decide whether a specific configuration of blocks is stable or not (see also \citep{baillargeon2002acquisition, baillargeon2004infants}). With $3$ months, infants decide based on a simple contact or no contact rule. According to this rule, a block configuration is considered to be stable if the two blocks touch each other. At around $5$ months, infants understand that the type of contact matters. Now, only configurations with blocks stacked on top of each other are judged as stable. At $6.5$ months, they additionally consider the overlap to determine the stability of a block configuration. Finally, at $12.5$ months they are able to incorporate the block shapes into their judgement, relying not only on the amount of contact but also on how the mass is distributed for each block. 
        
\begin{figure}[!h]
            \centering
            \includegraphics[width=\textwidth]{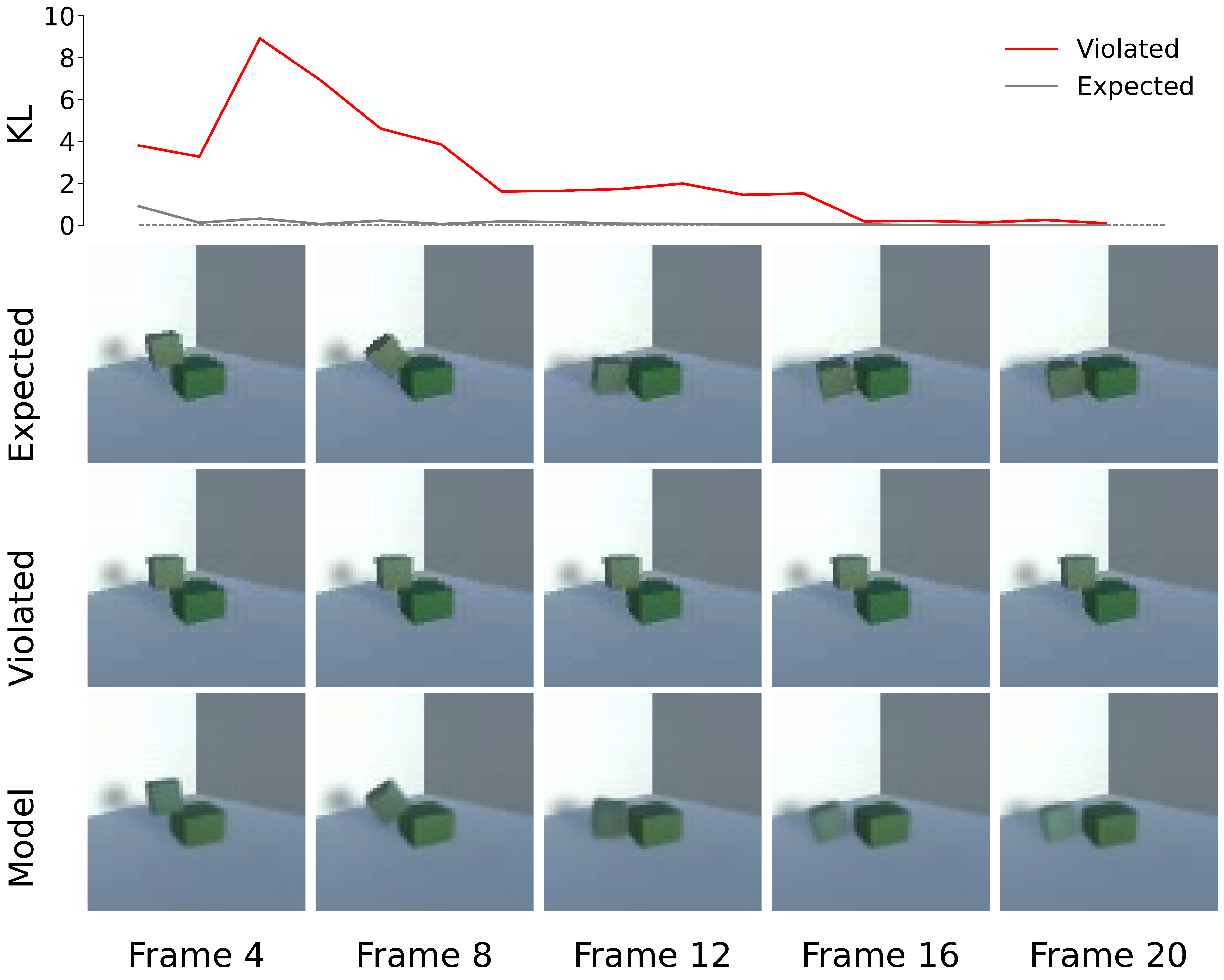}
            \caption{The first row shows the surprise for the expected and violated test sequences of the overlap rule. The second row shows the expected test sequence. The third row shows the violated test sequence. The last row shows the open-loop predictions from the model, given the first two frames of the violated test sequence. It is important to note that the first three frames were removed for this plot, as the uncertainty is very high when the model is first given the sequence.}
            \label{fig:kl_over_frames}
        \end{figure}

        To assess the learning trajectories of artificial systems, we generated a data set containing $100.000$ video sequences of support events using the Unity game engine \citep{unity}. The video sequences show stacks of two coloured blocks in a gray room and they consist of $20$ frames with a size of $64\times64$ pixels (see Figure \ref{fig:kl_over_frames}). We randomly varied a number of properties to ensure sufficient variability in the training data (see Appendix \ref{sec:stim_var} for further details). 
    
    \subsection{Modeling developmental trajectories}
    \label{sec:model}
    
        We investigate the development as stochastic optimization hypothesis by training a generative video prediction model using gradient descent. To obtain a learning trajectory of this model, we evaluate a snapshot of it in every epoch.
    
        We use the recurrent state space model (RSSM) \citep{hafner2019learning, saxena2021clockwork} as an exemplary model for our analysis. The RSSM can be seen as a sequential version of a variational autoencoder (VAE). It maintains a latent state at each time step, which is comprised of a deterministic component $h_t$ and a stochastic component $s_t$. These components depend on the previous time steps through a function $f(h_{t-1}, s_{t-1})$, which is implemented as a gated recurrent neural network. It is trained by optimizing the following evidence lower bound:
        \begin{align} 
        -\sum_{t=1}^T \mathbb{E}_{q(s_t\mid o_{\leq t})}[\textrm{ln } p(o_t\mid s_t)] + \mathbb{ E}_{q(s_{t-1}\mid o_{\leq t-1})}\big[\textrm{KL} (q(s_t \mid o_{\leq t}) \mid \mid p(s_t\mid s_{t-1}))\big] 
        \end{align}
        \label{eq:rssm_loss}
        
        After training, the RSSM can be used to generate open-loop predictions. For this, the model processes a number of initial observations to infer an approximate posterior $q(s_{t-1}\mid o_{\leq t-1})$. The subsequent open-loop predictions then result by decoding latent representations sampled from the prior $p(s_t\mid s_{t-1})$. We refer the reader to Appendix \ref{sec:model_imp} for further details about the model architecture and training procedure. 
        
    \subsection{Measuring surprise}
    \label{sec:measuring_surprise}
    
        In order to assess whether our model has understood a specific rule, we use the violation of expectation paradigm \citep{piloto2018probing, piloto2022learn}. The model is presented with two video sequences: a violated sequence, which constitutes a violation according to the rule, and an expected sequence, which is consistent with the rule. If the model has successfully learned a specific rule, it should show a larger degree of surprise for the violated compared to the expected sequence. Following \citet{piloto2018probing}, we measure the model's surprise using the Kullback–Leibler (KL) divergence between the prior and posterior over the latent representation \citep{baldi2010bits}. This approach closely resembles how developmental psychologists assess children's understanding of physical rules.
        
        We constructed a violated and an expected test sequence with identical image statistics for each of the four rules for support events outlined by \citet{baillargeon1996infants}. For example, according to the overlap rule, a block configuration should only be stable if the blocks are stacked on top of each other with enough overlap. The two test sequences for this rule therefore show two blocks that only slightly overlap (see Figure \ref{fig:kl_over_frames}). In the violated test sequence, the block configuration nonetheless appears stable, with the top block remaining on top of the bottom block. In contrast and consistent with the rule, the expected test sequence shows the top block falling. 
        
        \begin{figure}[!t]
            \centering
            \includegraphics[width=\textwidth]{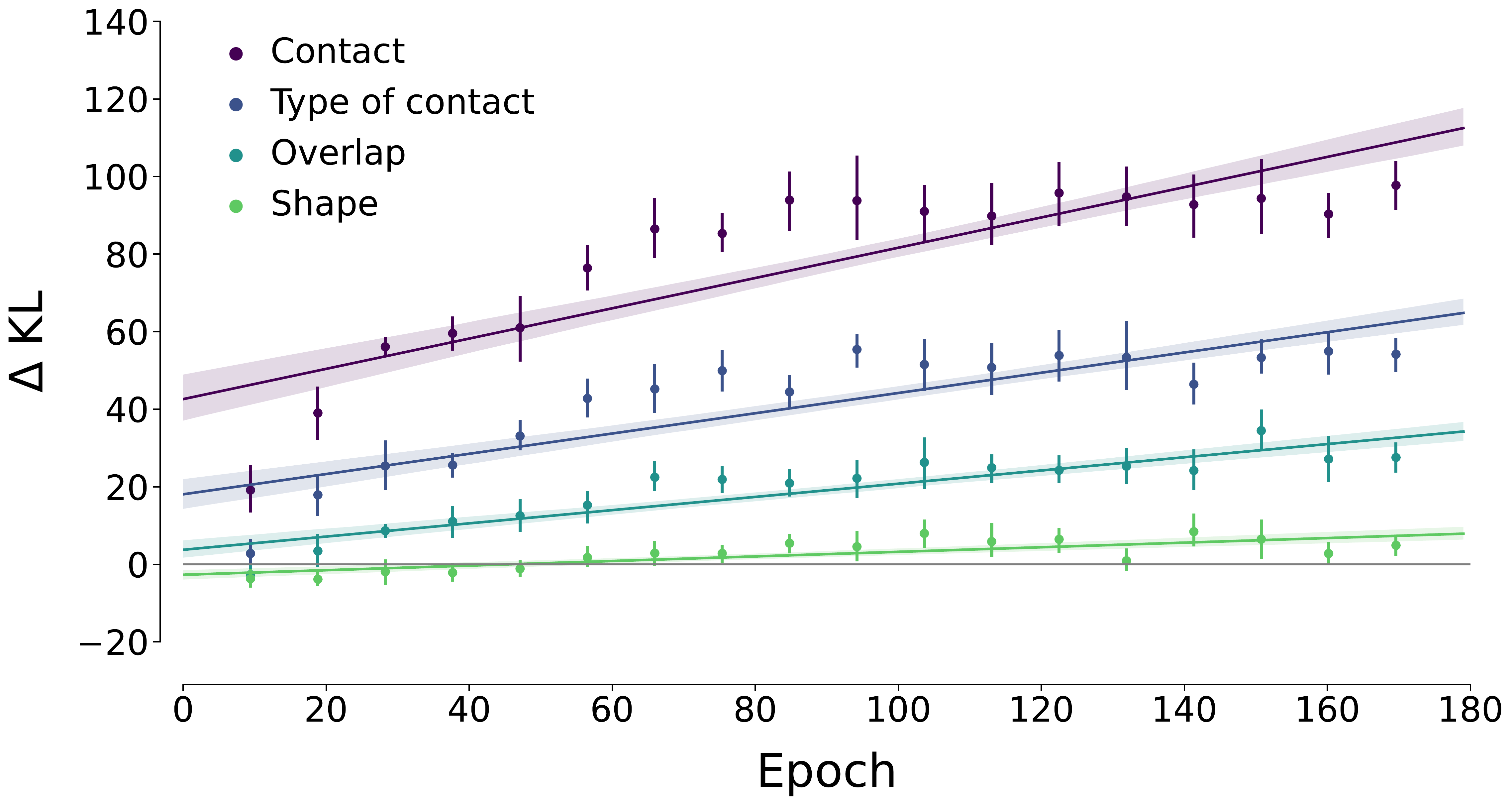}
            \caption{Linear regression fits for the difference between the surprise for the expected and the violated sequence at every epoch and for each of the four physical principles separately, as well as a 95\% confidence interval around the regression line. Additionally, the individual data points are binned in bins of size 10 and displayed with a 95\% confidence interval.}
            \label{fig:kl_over_epochs}
        \end{figure}
        
\section{Results}
\label{sec:results}

    Before investigating the learning trajectory of our model, we verified that it is able to predict the given scenes accurately into the future. For this purpose, we plotted the open-loop predictions of the fully trained model for all test sequences. Figure \ref{fig:kl_over_frames} illustrates the result for the overlap test sequences. We see that the predictions of the fully trained model closely match the expected sequence, indicating that the model has learned the overlap rule by the end of its training. This result also holds for the three other physical rules, see Appendix \ref{sec:figures} for the corresponding plots.   
    
    Next, we validated that the fully trained model is surprised when it observes a violation of a physical rule it has acquired. Figure \ref{fig:kl_over_frames} shows the KL divergence over the frames for the overlap test sequences. We see that the KL divergence for the violated sequence exceeds that for the expected sequence from around the fifth frame, which is when the two sequences diverge. 
    
    To quantitatively assess whether the model has learned the four physical rules, we computed the difference in KL divergence between the expected and violated sequences summed up over all frames. Figure \ref{fig:kl_over_epochs} shows this measure for each of the four physical rules over the course of training. We find that the difference in KL divergence is positive for all four physical rules in the fully trained model, confirming that it has learned all of them by the end of training.
    
    Finally, we investigated whether the stochastic optimization hypothesis can capture the developmental trajectories of children. For this purpose, we fitted a simple linear regression for each of the rules, using the difference in KL divergence as the dependent and training epochs as the independent variable. This directly relates to the stochastic optimization hypothesis, as the model becomes increasingly optimized over the epochs. The resulting regression coefficients were $0.390 \pm 0.024$, $0.261 \pm 0.017$, $0.170 \pm 0.011$, and $0.059 \pm 0.007$ for the contact or no contact, type of contact, overlap, and shape rules, respectively (all $p < 0.001$). We can see that the coefficients become smaller as the rules increase in complexity. The model first acquires the contact or no contact, followed by the type of contact, then the overlap, and finally the shape rule. This order of acquisition matches how children acquire these rules, thereby providing support to the idea of development as stochastic optimization.
    
\section{Discussion}
\label{sec:discussion}

    We have compared the learning trajectories of an artificial system to the developmental trajectories of children in the domain of physical reasoning. More specifically, we examined the idea of development as stochastic optimization. For this purpose, we used the violation of expectation paradigm to probe the knowledge of a modern deep generative neural network at different stages of its training process. We found that the model's learning trajectory resembled the trajectories of children during their cognitive development.
    
    In contrast to previous work in this domain \citep{binz2019emulating, giron2022developmental}, our modeling approach employs high-dimensional visual stimuli (i.e., video sequences) and solely relies on an unsupervised training objective. It, therefore, more closely mirrors the actual learning processes of children in the real world. Furthermore, the conducted experiments build on a well-established paradigm from developmental psychology. Together, these factors provide a high validity to our results. 
    
    Presently, the biggest mismatch between learning in our models and learning in children is that the latter do not observe a large number of support events. Instead, they simply witness the real world and generalize their acquired knowledge to the given experimental setting. To capture this process, we should ideally train our models in a similar way. This could, for example, be accomplished by utilizing the SAYCam data set, which contains a large number of longitudinal video recordings from infants’ perspectives \citep{sullivan2021saycam}. Additionally, this data set includes time stamps indicating when a child has encountered a particular scene, which could be used to investigate how the nature of the training data influences development.

    It also seems plausible that factors beyond stochastic optimization drive human development. There is, for example, evidence suggesting that children gain access to additional computational resources during their development, allowing them to apply more complex strategies \citep{binz2022exploration}. This idea of \emph{development as complexity increase} could be readily incorporated in our framework by replacing the evidence lower bound from Equation \ref{eq:rssm_loss} with a $\beta$-VAE objective \citep{higgins2016beta, burgess2018understanding}. 

    To fully disentangle these hypotheses, a single paradigm will likely not be sufficient. Instead, we will need to go beyond support events and investigate developmental trajectories across a variety of different tasks. Findings that hold across domains would further increase the reliability of our results and have potential implications for cognitive psychology and artificial intelligence alike.

\begin{ack}
This work was funded by the Max Planck Society and the Volkswagen Foundation. 
\end{ack}

\clearpage
\bibliographystyle{svrhm_2022}
\bibliography{svrhm_2022}

\clearpage
\section*{Checklist}


\begin{enumerate}

    \item For all authors...
    \begin{enumerate}
      \item Do the main claims made in the abstract and introduction accurately reflect the paper's contributions and scope?
        \answerYes{Our main claims accurately reflect the paper's contributions.}
      \item Did you describe the limitations of your work?
        \answerYes{We have described the limitations of our work, see section \ref{sec:discussion}.}
      \item Did you discuss any potential negative societal impacts of your work?
        \answerNo{We do not think that there are negative societal impacts as a result of this work.}
      \item Have you read the ethics review guidelines and ensured that your paper conforms to them?
        \answerYes{We have read the ethics review guidelines.}
    \end{enumerate}

    \item If you are including theoretical results...
    \begin{enumerate}
      \item Did you state the full set of assumptions of all theoretical results?
        \answerNA{We do not include theoretical results.}
            \item Did you include complete proofs of all theoretical results?
        \answerNA{We do not include theoretical results.}
    \end{enumerate}

    \item If you ran experiments...
    \begin{enumerate}
      \item Did you include the code, data, and instructions needed to reproduce the main experimental results (either in the supplemental material or as a URL)?
        \answerNo{However, the code, data, and instructions are available upon request.}
      \item Did you specify all the training details (e.g., data splits, hyperparameters, how they were chosen)?
        \answerYes{We have specified the training details, see sections \ref{sec:model} and \ref{sec:model_imp}.}
    \item Did you report error bars (e.g., with respect to the random seed after running experiments multiple times)?
        \answerYes{We report error bars for the regression analysis (see section \ref{sec:results}) and in figure \ref{fig:kl_over_epochs}.}
    \item Did you include the total amount of compute and the type of resources used (e.g., type of GPUs, internal cluster, or cloud provider)?
        \answerNo{We did not track the total amount of compute used for testing. However, we have mentioned the training time and GPU used, see section \ref{sec:model_imp}.}
    \end{enumerate}

    \item If you are using existing assets (e.g., code, data, models) or curating/releasing new assets...
    \begin{enumerate}
      \item If your work uses existing assets, did you cite the creators?
        \answerYes{Our model implementation builds on a previous implementation, which we cited in section \ref{sec:model_imp}.}
      \item Did you mention the license of the assets?
        \answerNo{We did not mention the licence in the manuscript, however the previous implementation is licensed under the MIT License.}
      \item Did you include any new assets either in the supplemental material or as a URL?
        \answerNo{However, the data set we created is available upon request.}
      \item Did you discuss whether and how consent was obtained from people whose data you're using/curating?
        \answerNA{We did not use or curate data from other people.}
      \item Did you discuss whether the data you are using/curating contains personally identifiable information or offensive content?
        \answerNA{We did not use or curate data from other people.}
    \end{enumerate}

    \item If you used crowdsourcing or conducted research with human subjects...
    \begin{enumerate}
      \item Did you include the full text of instructions given to participants and screenshots, if applicable?
        \answerNA{We did not use crowdsourcing or conduct research with human subjects.}
      \item Did you describe any potential participant risks, with links to Institutional Review Board (IRB) approvals, if applicable?
        \answerNA{We did not use crowdsourcing or conduct research with human subjects.}
      \item Did you include the estimated hourly wage paid to participants and the total amount spent on participant compensation?
        \answerNA{We did not use crowdsourcing or conduct research with human subjects.}
    \end{enumerate}

\end{enumerate}


\appendix

\section{Appendix}

    \subsection{Stimulus variations}
    \label{sec:stim_var}
        
        The following variables were randomly varied in order to ensure sufficient variability in the data set: lower block size, lower block color, upper block color, lower block rotation, upper block rotation, upper block position (offset) and camera angle. Additionally, the shape of the upper block was varied: half of the trials featured a cube as an upper block, while the other half featured an L-shaped block with randomly sampled side lengths.
        
    \subsection{Model implementation}
    \label{sec:model_imp}
    
        The recurrent state space model (RSSM) can be seen as a sequential VAE \citep{hafner2019learning, saxena2021clockwork}. The deterministic component $h_t$ and stochastic component $s_t$ depend on the deterministic and stochastic components at the previous time steps through $f(h_{t-1}, s_{t-1})$, which is implemented through a gated recurrent unit cell. As outlined by \citet{hafner2019learning}, the individual components of the RSSM are:

        \begin{align*} 
        &\text{Deterministic state model:}  &   h_t &= f(h_{t-1}, s_{t-1}) \\
        &\text{Stochastic state model:}     &   s_t &\sim p(s_t \mid h_t) \\ 
        &\text{Observation model:}          &   o_t &\sim p(o_t \mid h_t, s_t) \\
        \end{align*}
        
        The models were implemented in PyTorch \citep{paszke2019pytorch}. For all models, the size of the stochastic hidden dimension $s_t$ was kept at 20, while the size of the deterministic hidden dimension $h_t$ was set to 200, as in previous implementations of the RSSM \citep{hafner2019learning, saxena2021clockwork}. 
        
        We used the encoder and decoder from \citet{dittadi2020transfer}. The encoder consists of three blocks. The first block consists of a convolutional layer with a kernel of size 5 and a stride of 2 and a padding of 2, followed by a leaky ReLU activation function, followed by two residual blocks. The second block consists of a convolutional layer with a kernel of size 1 and a stride of 1 and no padding, followed by average pooling with a kernel of size 2, followed by two blocks residual blocks. The third block consists of average pooling with a kernel of size 2, followed by two blocks residual blocks. The fourth block consists of a convolutional layer with a kernel of size 1 and a stride of 1 and no padding, followed by average pooling with a kernel of size 2, followed by two blocks residual blocks. The fifth block consists of average pooling with a kernel of size 2, followed by two blocks residual blocks.
        
        The decoder consists of five blocks. The first block consists of two residual blocks, followed by upsampling with a scale factor of 2. The second block consists of two residual blocks, followed by a deconvolutional layer with a kernel size of 1 and a stride of 1, followed by upsampling with a scale factor of 2. The third block again consists of two residual blocks, followed by upsampling with a scale factor of 2. The fourth block consists of two residual blocks, followed by a deconvolutional layer with a kernel size of 1 and a stride of 1, followed by upsampling with a scale factor of 2. The fifth block consists of two residual blocks, followed by upsampling with a scale factor of 2, a leaky ReLU activation funktion, followed by a deconvolutional layer with a kernel size of 5 and a stride of 1 and a padding of 2.
        
        The models were trained for 200 epochs using a batch size of 32. The stimulus sets were randomly split into 99.000 training sequences and 1000 validation sequences. The loss function was optimized using the Adam optimiser with a learning rate of 0.001 \citep{kingma2014adam}, which was divided by 10 every 50 epochs. The models were trained on a NVIDIA Quadro RTX 5000 for roughly 7 days. Our implementation of the RSSM borrows from a previous implementation on \href{https://github.com/cross32768/PlaNet_PyTorch}{GitHub}. The complete code for this project, including our model implementation, is available upon request.

    \clearpage
    \subsection{Figures}
    \label{sec:figures}
    \vspace{1cm}

    \begin{figure}[!h]
        \centering
        \includegraphics[width=\textwidth]{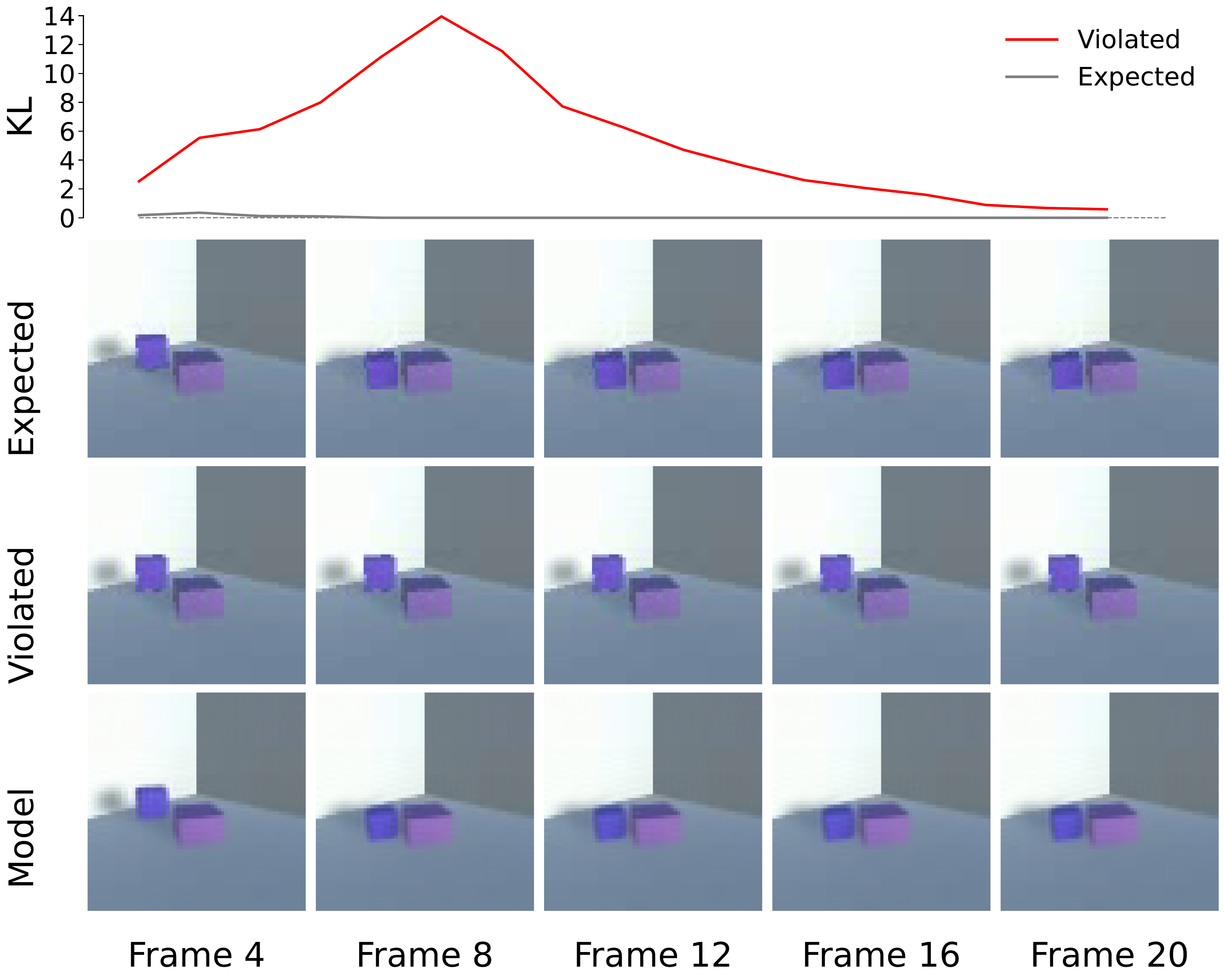}
        \caption{The first row shows the surprise for the expected and violated test sequences of the contact or no contact rule. The second row shows the expected test sequence. The third row shows the violated test sequence. The last row shows the open-loop predictions from the model, given the first two frames of the violated test sequence. It is important to note that the first three frames were removed for this plot, as the uncertainty is very high when the model is first given the sequence.}
    \end{figure}
    
    \begin{figure}[!t]
        \centering
        \includegraphics[width=\textwidth]{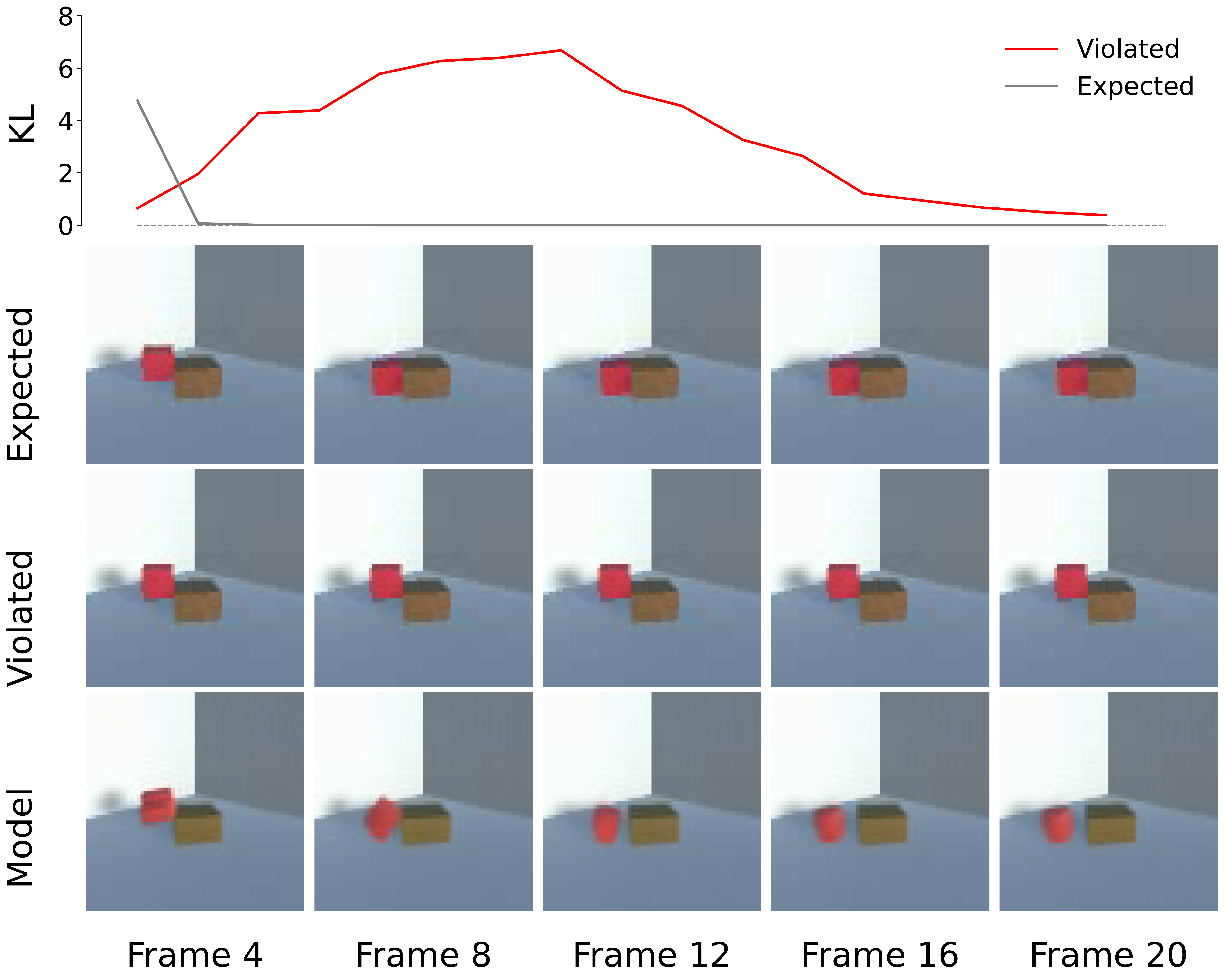}
        \caption{The first row shows the surprise for the expected and violated test sequences of the type of contact rule. The second row shows the expected test sequence. The third row shows the violated test sequence. The last row shows the open-loop predictions from the model, given the first two frames of the violated test sequence. It is important to note that the first three frames were removed for this plot, as the uncertainty is very high when the model is first given the sequence.}
    \end{figure}
    
    \begin{figure}[!t]
        \centering
        \includegraphics[width=\textwidth]{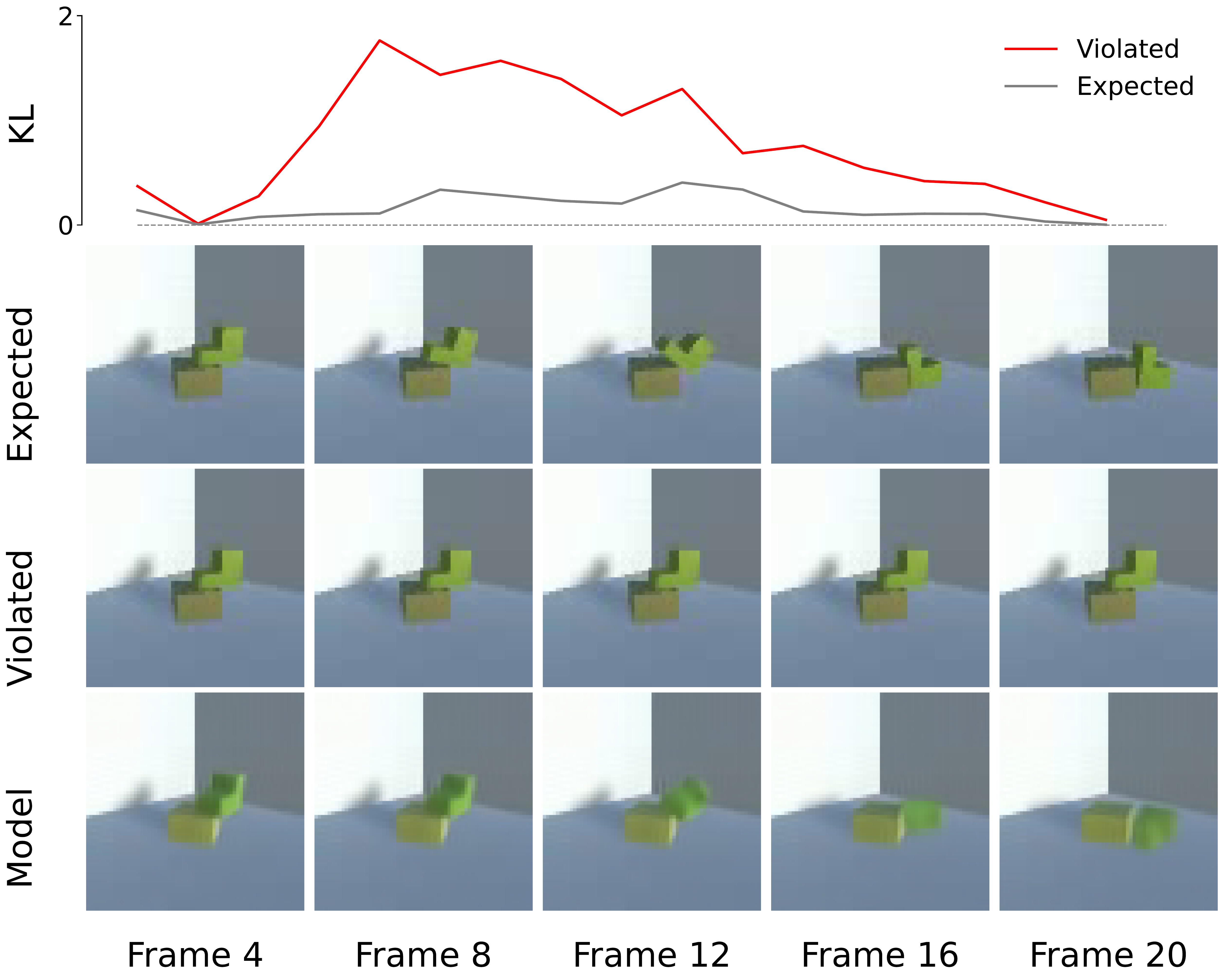}
        \caption{The first row shows the surprise for the expected and violated test sequences of the shape rule. The second row shows the expected test sequence. The third row shows the violated test sequence. The last row shows the open-loop predictions from the model, given the first two frames of the violated test sequence. It is important to note that the first three frames were removed for this plot, as the uncertainty is very high when the model is first given the sequence.}
    \end{figure}
    
\end{document}

%% file: math_commands.tex
\usepackage{amsmath,amsfonts,bm}

\def\eqref#1{equation~\ref{#1}}

\def\1{\bm{1}}

\DeclareMathAlphabet{\mathsfit}{\encodingdefault}{\sfdefault}{m}{sl}
\SetMathAlphabet{\mathsfit}{bold}{\encodingdefault}{\sfdefault}{bx}{n}

